\documentclass[letterpaper]{article} 
\usepackage{aaai25}  
\nocopyright 

\usepackage{times}  
\usepackage{helvet}  
\usepackage{courier}  
\usepackage[hyphens]{url}  
\usepackage{graphicx} 
\usepackage{natbib}  
\usepackage{caption} 
\usepackage{algorithm}
\usepackage{algorithmic}

\usepackage{newfloat}
\usepackage{listings}
\DeclareCaptionStyle{ruled}{labelfont=normalfont,labelsep=colon,strut=off} 
\floatstyle{ruled}
\newfloat{listing}{H}{lst}{}
\floatname{listing}{Listing}
\usepackage{amsmath}
\usepackage{booktabs}

\title{MedDiT: A Knowledge-Controlled Diffusion Transformer Framework for Dynamic Medical Image Generation in Virtual Simulated Patient}

\author{
    Yanzeng Li\textsuperscript{\rm 1,2},
    Cheng Zeng\textsuperscript{\rm 3},
    Jinchao Zhang\textsuperscript{\rm 2}\thanks{Coresponding Authors},
    Jie Zhou\textsuperscript{\rm 2},
    Lei Zou\textsuperscript{\rm 1*}
}
\affiliations{
    \textsuperscript{\rm 1}Wangxuan Institute of Computer Technology, Peking University\\
    \textsuperscript{\rm 2}Pattern Recognition Center, WeChat AI, Tencent Inc, China\\
    \textsuperscript{\rm 3}School of Computer Science, Wuhan University\\

    liyanzeng@stu.pku.edu.cn, zengc@whu.edu.cn, zoulei@pku.edu.cn \\
    \{dayerzhang, withtomzhou\}@tencent.com, 
}

\begin{document}

\maketitle

\begin{abstract}
Medical education relies heavily on Simulated Patients (SPs) to provide a safe environment for students to practice clinical skills, including medical image analysis. However, the high cost of recruiting qualified SPs and the lack of diverse medical imaging datasets have presented significant challenges. To address these issues, this paper introduces MedDiT, a novel knowledge-controlled conversational framework that can dynamically generate plausible medical images aligned with simulated patient symptoms, enabling diverse diagnostic skill training. Specifically, MedDiT integrates various patient Knowledge Graphs (KGs), which describe the attributes and symptoms of patients, to dynamically prompt Large Language Models' (LLMs) behavior and control the patient characteristics, mitigating hallucination during medical conversation. Additionally, a well-tuned Diffusion Transformer (DiT) model is incorporated to generate medical images according to the specified patient attributes in the KG. In this paper, we present the capabilities of MedDiT through a practical demonstration, showcasing its ability to act in diverse simulated patient cases and generate the corresponding medical images. This can provide an abundant and interactive learning experience for students, advancing medical education by offering an immersive simulation platform for future healthcare professionals. The work sheds light on the feasibility of incorporating advanced technologies like LLM, KG, and DiT in education applications, highlighting their potential to address the challenges faced in simulated patient-based medical education.
\end{abstract}

%

\section{Introduction}

Medical education plays a crucial role in preparing future healthcare professionals, relying extensively on Simulated Patients (SPs) to provide a safe and controlled environment for practicing clinical skills~\cite{gaba2007future,ziv2006simulation,sanko2013establishing,mesquita2010developing}. However, the traditional use of SPs presents significant challenges, primarily due to the high costs associated with recruiting and training qualified individuals~\cite{hillier2020standardization, Felix2019TypesOS}. While the development of Large Language Models (LLMs) offers the potential to build practical virtual SPs (VSPs)~\cite{chen2023llm, benitez2024harnessing, holderried2024generative, li2024leveraging}, the scarcity of diverse and comprehensive medical imaging datasets complicates the ability of VSPs to provide varied and realistic training scenarios~\cite{glatard2012virtual, baraheem2023image, wang2021review}.

To address these limitations, we introduce MedDiT, a novel VSP framework designed to enhance the educational experience. The core of MedDiT's functionality is the integration of patient Knowledge Graphs (KGs)~\cite{fensel2020introduction, gyrard2018personalized}. These KGs meticulously describe patient attributes and symptoms, serving as a foundation to guide the behavior of LLMs. By dynamically prompting LLM behavior, MedDiT ensures that patient characteristics are accurately represented, effectively mitigating issues such as hallucination during medical conversations~\cite{li2024leveraging}.
In addition to its conversational capabilities, MedDiT incorporates a series of well-tuned Diffusion Transformer (DiT) models~\cite{yang2023diffusion, pan20232d}. These models can generate medical images that correspond to the specified patient attributes within the KGs, providing a realistic and varied set of scenarios for students to engage with.

Essentially, MedDiT is a multi-agent system that integrates KG agent, chat agent, and image generation agent. This integration is centered around KG data, facilitating interaction and enabling the system to dynamically generate medical images that align with simulated patient symptoms. Through a practical demonstration, we showcase MedDiT's ability to simulate diverse patient cases and generate the corresponding medical images. This not only enhances the learning experience by offering an abundant and interactive platform but also advances medical education by providing an immersive simulation environment.

\section{System Design \& Architecture}

\begin{figure*}[ht]
    \centering{\includegraphics[width=\linewidth]{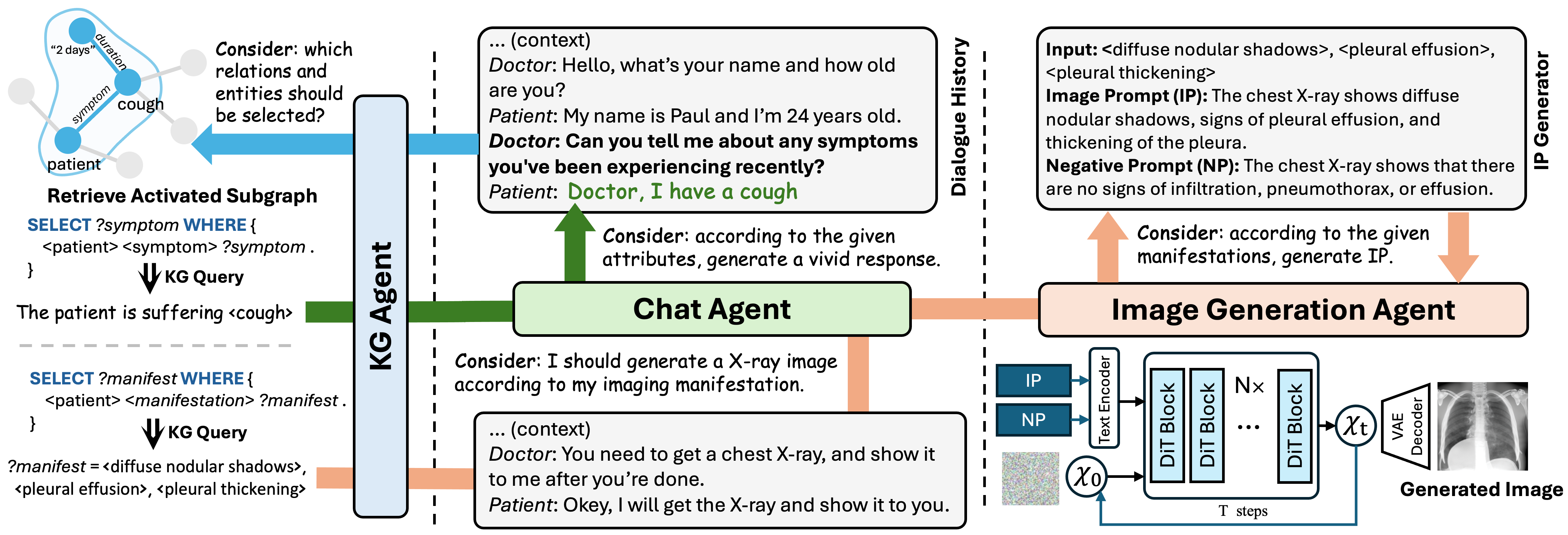}}
        \caption{The overview diagram of MedDiT. There are 3 distinct LLM-based agents for controlling information flow across various modalities, including graph, text and image.}
        \label{fig:overall}
    \end{figure*}

\begin{figure}[t]
    \centering
    \includegraphics[width=\linewidth]{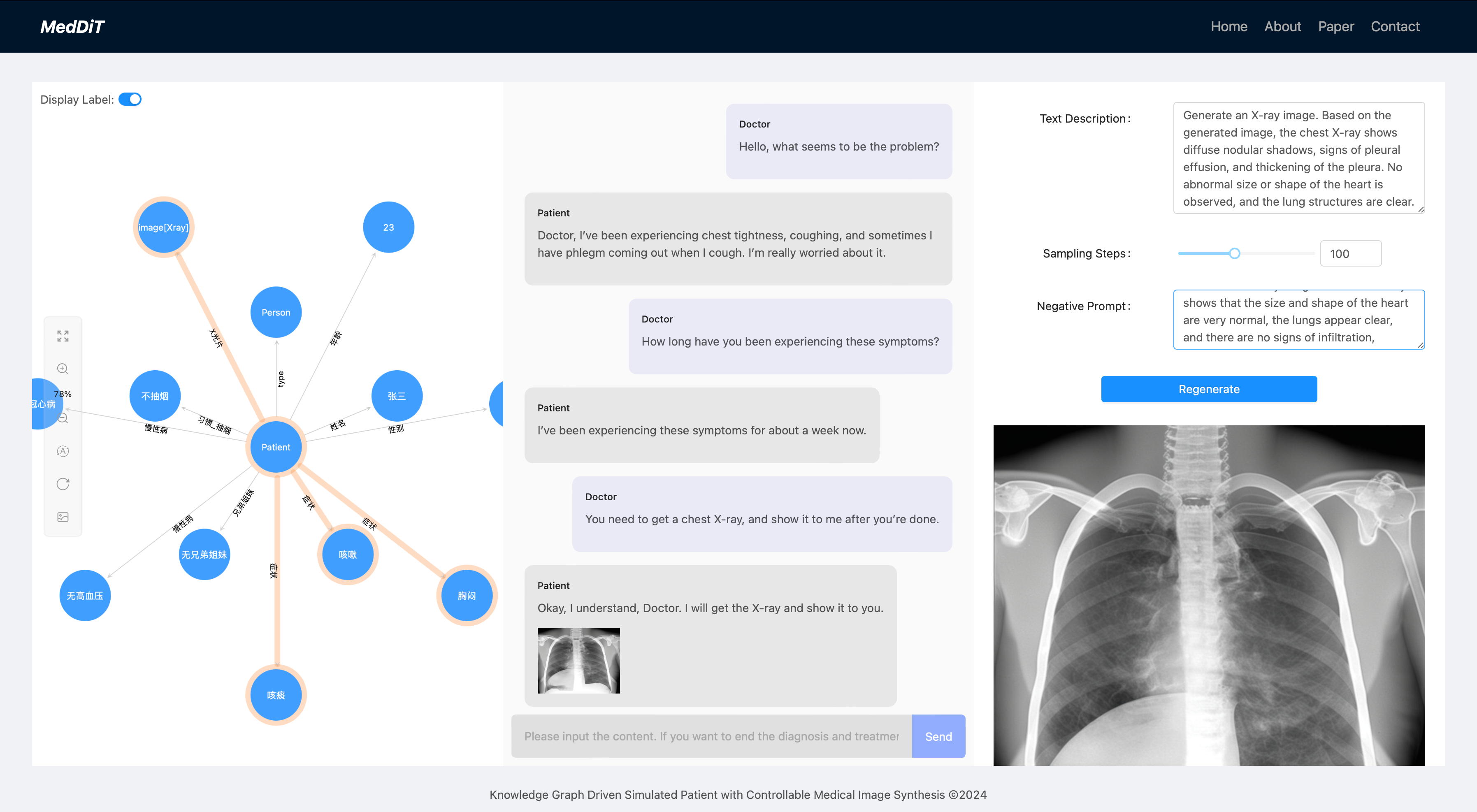}
    \caption{
    Screenshot of MedDiT. Left: Visualization of patient KG, indicating the activated subgraph for prompting conversaion. Center: The dialogue interface. Right: The DiT model interface. 
    }
    \label{fig:screenshot}
\end{figure}

We developed MedDiT with a focus on modularity and configurability. The KG, chat, and multimodal components are constructed as microservices, while a unified LLM server is utilized to support the entire prompt workflow. In practice, we employ the Qwen2 72B instruction-tuned version~\cite{qwen2}\footnote{\url{https://huggingface.co/Qwen/Qwen2-72B-Instruct}} as our backbone LLM for building all the agents. The overview of MedDiT is illustrated in Figure~\ref{fig:overall}. Figure~\ref{fig:screenshot} presents demonstration of MedDiT.

\paragraph{KG-enhanced LLM-based VSP}

To construct a controllable and less-hallucinatory VSP system, we introduce the patient KG as the core information source as~\citet{li2024leveraging}. A KG agent is applied to retrieve the conversation-related subgraph by discerning the user's intention and generating a SPARQL query for extracting related entities, relations, and attributes in KG. Subsequently, the extracted subgraph is linearized into natural language to serve as role-setting for prompting the chat agent's responses. In this manner, MedDiT maintains dialogue consistency effectively, reduces the hallucinations, and saves token costs.

\paragraph{KG-controlled DiT Model}

We utilize HunyuanDiT~\cite{li2024hunyuandit} as our backbone generative model and train a LoRA adaptor for the transfer model to adapt to our target domain~\cite{hu2021lora}. In this demonstration, we selected a subset of Chest X-ray images from the Open-i dataset~\cite{demner2016preparing} for training, comprising 3,314 images along with their corresponding textual descriptions. Detailed training parameters are provided in Appendix~\ref{sec:hyperparameters}. Subsequently, we construct an agent to generate image prompts from the structured manifestations, which are stored in the KG and retrievable by KG agent. Once generated, the images are displayed in dialogue flow for students' further analysis.

\paragraph{SP Evaluation} As a VSP system, it is crucial to assess the dialogue history between the SP and students, score the interactions, and provide feedback to help students enhance their medical conversation skills through iterative practice. Following the instructions from \citet{li2024leveraging}, we utilize various prompts and auxiliary models to analysis the dialogue history across multiple aspects, including the completeness of necessary information, thoroughness of symptom inquiry, and the emotions conveyed to the patient. Finally, a detailed evaluation report, including scoring and advice, is generated by the LLM based on these criteria and the gold standard for the corresponding VSP case. Appendix~\ref{sec:assess-exp} presents an actual evaluation report.

\section{Conclusion and Future Work}

In conclusion, MedDiT provides a comprehensive and practical solution for building VSP systems that can dynamically generate medical images. By developing a multi-agent system based on LLMs, MedDiT has successfully integrated KGs to control the chat flow and utilized DiT models to produce medical images that align with the context and symptoms of patients. The system's ability to generate diverse and realistic medical images, combined with its interactive conversational capabilities, offers an abundant learning experience that can significantly improve medical education.
In future work, we plan to further expand the scope of the KG to encompass a broader range of medical conditions, symptoms, and patient profiles. We will also increase the number of SP cases and explore training more kinds of medical image adaptors. Furthermore, we aim to conduct evaluation experiments on MedDiT to test large vision models and investigate the possibility of using large multi-modal models for comprehensive diagnosis.

\bibliography{aaai25}

\appendix

\section{Knowledge Controlled Image Generation}

In this study, we propose a framework for generating knowledge-driven imagery from structured KGs, delineated through following processes:

A KG can be denoted as $G = (E,R,T)$, where $E$ represents entities, $R$ is the set of relations, and $T \subseteq E \times R \times E$ is the collection of knowledge triples $\langle s,p,o \rangle$, with $s,o \in E, p\in R$ being subjects, objects, and predicates, respectively.
We define a function that represents the knowledge implied by arguments as: $$\text{Knowledge}: X \rightarrow \mathcal{K},$$ where $X$ can be a graph, text, or images, and $\mathcal{K}$ is the set of knowledge representations, implying that knowledge can be represented by $\langle s,p,o \rangle$.

Initially, we perform subgraph retrievel, where a subgraph $G' = (E', R', T')$ is derived from the original patient graph $G = (E, R, T)$, ensuring that $E' \subseteq E$, $R' \subseteq R$, $T' \subseteq T$, and $\text{Knowledge}(G') \subseteq \text{Knowledge}(G)$. This subgraph serves as the foundational data structure for subsequent operations. The retrieved subgraph $G'$ can be represented by: $$ f: G' \rightarrow \text{Text}, \text{s.t.} \quad \text{Knowledge}(\text{Text})\approx \text{Knowledge}(G'),$$ which maps the subgraph to a textual representation while preserving the inherent knowledge, such that the knowledge contained in the text equals that of the subgraph. In practice, we employ a strong LLM as the $f$ for encoding subgraph to text as losslessly as possible. 
Subsequently, this textual representation $\text{IP} = f(G')$ is utilized as an image prompt within the DiT model, denoted as $$ I = \text{DiT}(\text{IP}),$$ and the total progress can be denoted as: $$g: \text{Text} \rightarrow I, \text{s.t.} \quad I \models \text{Knowledge}(G'),$$ ensuring that the generated image $I$ accurately reflects the knowledge embedded in $G'$. This comprehensive approach $g$ describes the integration of graph-based data with text and image modalities, fostering enhanced interpretability and control in image synthesis.

\section{Hyperparameter Settings}~\label{sec:hyperparameters}

The hyperparameters used to train the DiT model for generating responses during medical conversations are displayed in Table \ref{tab:hyper}. The unmentioned parameters are aligned with the configuration of~\citet{li2024hunyuandit}.

\begin{table}[h]
\centering
\resizebox{1.\columnwidth}{!}{
\begin{tabular}{c l c}
    \toprule
    \bfseries Name & \bfseries Description & \bfseries Setting\\
    \cmidrule(lr){1-3}
    $size$ & Image size & 1024 $\times$ 1024 \\
    $d_\text{rank}$ & LoRA rank & 64\\
    $t_\text{lora}$ & Weights applied with LoRA & $W_q, W_k, W_v, W_\text{out}$ \\
    $d_\text{text}$ & Hidden size of the T5 encoder & 2048 \\
    $d_\text{clip}$ & Hidden size of the CLIP encoder & 1024 \\
    $l_\text{text}$ & Maximum length of the T5 encoder & 256 \\
    $lr$ & Learning rate & 1e-4 \\
    $epoch$ & Training epochs & 100\\
    $optim$ & Optimizer for training LoRA adaptor & Adam\\
    \bottomrule
\end{tabular}}
\caption{Hyperparameter settings for our approach.}\label{tab:hyper}
\end{table}

\section{Assessment Example}\label{sec:assess-exp}

In Listing \ref{lst:example}, we present an example of a generated assessment designed to evaluate a student's performance. MedDiT concludes with a comprehensive score and a letter of advice to guide the student in their future practice.

\begin{listing}
\caption{An assessment example of a student's performance in medical conversation.}
\label{lst:example}
\begin{lstlisting}
The ID for this assessment is {{ID}}, and your score is 69/100 points. Here are some suggestions for you:

### Summary:
During the process of assessing patients' health conditions, you have paid attention to many key factors, such as marriage, smoking, infectious diseases, vomiting, hepatitis, fatigue, tuberculosis, and parents' information. However, you seem to have not fully covered some important details, such as genetic history and symptoms.
As a medical professional, your work is extremely important and indispensable. Your focus on existing fields is commendable. Improving comprehensiveness, not only focusing on current symptoms but also delving into patients' past medical history and family health conditions, can help you develop more targeted treatment plans and anticipate potential health problems.

### Improvement Guidance:
Genetic History: You can briefly inquire about the patient's family medical history, especially those related to genetics, such as cardiovascular diseases, diabetes, cancer, etc. This helps with prevention and intervention efforts and early identification of high-risk groups.
Symptom Assessment: In every consultation, it is essential to fully understand the patient's symptoms. This includes not only current symptoms but also any discomfort experienced in the past few months or years. This helps to piece together a complete health picture and identify clues that may require further examination.
Comprehensiveness and Practicality: Your work is constantly committed to expanding knowledge boundaries and improving clinical skills. Encourage you to continue exploring the latest medical research, which not only helps you acquire new information but also provides strong support when dealing with complex diseases.
On this basis, every effort you make demonstrates an increasing attention to and understanding of patients' health, which is a significant improvement. Maintain this critical thinking and continuous learning attitude, and your career will be substantially enhanced, bringing higher quality medical services to patients.
\end{lstlisting}
\end{listing}

\end{document}